# An Interval Type-2 Version of Bayes' Theorem Derived from Interval Probability Range Estimates Provided by Subject Matter Experts

John T. Rickard, *Senior Member, IEEE*, William A. Dembski, *Senior Member, IEEE,*

James Rickards*, Member, IEEE*

*Abstract*—Bayesian inference is widely used in many different fields to test hypotheses against observations. In most such applications, an assumption is made of precise input values to produce a precise output value. However, this is unrealistic for real-world applications. Often the best available information from subject matter experts (SMEs) in a given field is interval range estimates of the input probabilities involved in Bayes' Theorem. This paper provides two key contributions to extend Bayes' Theorem to an interval type-2 (IT2) version. First, we develop an IT2 version of Bayes' Theorem that uses a novel and conservative method to avoid potential inconsistencies in the input IT2 MFs that otherwise might produce invalid output results. We then describe a novel and flexible algorithm for encoding SME-provided intervals into IT2 fuzzy membership functions (MFs), which we can use to specify the input probabilities in Bayes' Theorem. Our algorithm generalizes and extends previous work on this problem that primarily addressed the encoding of intervals into word MFs for Computing with Words applications.

## I. INTRODUCTION

**B**ayes' Theorem (BT) [1] is one of the earliest and most fundamental results in all of probability theory. BT provides the mathematical formula for the *a posteriori* probability $P(H|E)$ of a hypothesis $H$, given an event $E$ that is typically an observation, as a function of the conditional probability $P(E|H)$ of the event's occurrence given $H$ (often denoted as the *likelihood function*) and the *a priori*

Submitted April 21, 2025.
John T. Rickard is with Royalty & Streaming Advisors LLC, Larkspur, CO, USA and Meraglim Holdings Corp. (email: terry.rickard@hushmail.com).
William A. Dembski is with Discovery Institute's Center for Science and Culture (email: billdembski@gmail.com).
James Rickards is with Royalty & Streaming Advisors LLC, Larkspur, CO, USA, and is the Editor of the financial newsletter *Strategic Intelligence*. (email: james.rickards@gmail.com).

probabilities $P(H)$ and $P(E)$ of $H$ and $E$, respectively:

$$P(H|E) = \frac{P(E|H)P(H)}{P(E)}. \qquad (1)$$

This basic relationship is the foundation for the extremely broad subject area of Bayesian inference (BI) [2]-[6]. BI is employed in numerous fields including medicine [7]-[8], search processes [10],[11], machine learning [12],[13], intelligence analysis [14],[15] and financial forecasting [16],[17], to name but a few among a vast corpus of potential references. Indeed, a web search on "Bayes' Theorem in XXX" where XXX represents one's scientific discipline of choice, will typically return numerous references.

Of course, in real-world applications, one seldom ever can know precisely the probabilities on the right-hand side of (1). This imprecise knowledge leads naturally to the notion of fuzzifying BT to enable a fuzzified form of BI. Previous work on fuzzifying BI (e.g., see [18]-[20]) typically begins with making parametric probability distributional assumptions regarding the likelihood and *a priori* probabilities in (1), assigning fuzzy membership functions (MFs) to the parameters of these distributions, and then calculating a corresponding fuzzified probability distribution for $P(H|E)$ via averaging over the fuzzy MFs.

In the present work, we take a more direct and subject matter expert (SME) informed approach to the problem that avoids having to make any distributional assumptions regarding the BT inputs. Instead, we begin with the solicitation of interval range estimates for each of the probabilities on the right-hand side of (1) from a group of SMEs in the domain and application of interest. We then construct corresponding interval type-2 (IT2) MFs from these interval sets using a novel algorithm that substantially generalizes previous work involving the encoding of interval range estimates into word representations [21]-[28]. This encoding algorithm for mapping interval sets to IT2 MFs represents one of the major contributions of this paper.

Once these IT2 MFs for the three input probabilities are constructed, we then perform a direct calculation of the corresponding IT2 MF for $P(H|E)$ using interval arithmetic operations upon the $\alpha$-cuts of the upper and lower MFs (UMF and LMF, respectively) of $P(E|H)$, $P(H)$ and $P(E)$. In some instances, the interval division operation is imperiled by the possible overlap of the denominator $\alpha$-cuts for $P(E)$ with those of the product $\alpha$-cuts for the numerator $P(E|H)P(H)$, which produces invalid results where the corresponding $\alpha$-cut intervals for $P(H|E)$ have interval values exceeding unity. We propose a conservative and analytically sound method for thwarting this potential problem, resulting in a feasible and intuitive IT2 MF for $P(H|E)$. This method represents the other major contribution of this paper.

The organization of this paper is as follows. Section II presents our IT2 version of Bayes' Theorem and illustrates it with examples. We present this material first since these results are independent of the way one obtains the input IT2 MFs. Section III presents a novel algorithm for synthesizing IT2 MFs from sets of interval range estimates supplied by SMEs. This algorithm can process input intervals pertaining to any physical or technical quantities, over both bounded and unbounded domains. Section IV illustrates this new algorithm with some examples. Section V concludes.

## II. TYPE-2 VERSION OF BAYES' THEOREM

Suppose we have obtained SME-supplied IT2 MFs for each of the probabilities $P(E|H)$, $P(H)$ and $P(E)$ appearing on the right-hand side of Bayes' Theorem in (1), e.g., by using the algorithm described in Sections III and IV. The numerator product $P(E|H)P(H)$ IT2 MF can then be calculated directly from interval multiplications of the corresponding $\alpha$-cuts of the UMF and LMF of $P(E|H)$ and $P(H)$.

One might then apply interval division to the corresponding $\alpha$-cuts of the numerator product IT2 UMF/LMF and those of the denominator IT2 UMF/LMF for $P(E)$. However, in cases where any $\alpha$-cuts of the latter MF overlaps with the corresponding $\alpha$-cuts of the numerator MF, this results in invalid $\alpha$-cut values for the IT2 UMF and/or LMF of $P(H|E)$, i.e., $\alpha$-cut intervals whose range exceeds unity. This problem arises when the sets of SME-provided interval range estimates for the probabilities $P(E|H)$, $P(H)$ and $P(E)$ may have mutual inconsistencies.

To overcome this difficulty, we first observe the inequality:
$$P(E) \geq P(E,H) = P(E|H)P(H). \quad (2)$$
The implication of this inequality for MFs is that, since any value of the probabilities within the support range of an $\alpha$-cut has non-zero membership, perhaps excluding its endpoints, *no $\alpha$-cut of the UMF or LMF of $P(E)$ is permitted to overlap with its corresponding $\alpha$-cut for the numerator $P(E|H)P(H)$ except possibly at an endpoint.*

To accommodate the inequality in (2) in a conservative fashion, we propose the following strategy. Let the respective $\alpha$-cut intervals of the BT (UMF/LMF) product $P(E|H)P(H)$ and those of $P(E)$ be denoted as $[PE_\ell(\alpha), PE_r(\alpha)]$ and $[E_\ell(\alpha), E_r(\alpha)]$, respectively. For every $\alpha$ such that $E_\ell(\alpha) < PE_r(\alpha)$, i.e., where a $P(E)$ UMF or LMF $\alpha$-cut overlaps with its corresponding $P(E|H)P(H)$ $\alpha$-cut, we replace the $P(E)$ $\alpha$-cut with the interval:
$$\left[\max\left(PE_r(\alpha), E_\ell(\alpha)\right), \max\left(PE_r(\alpha), E_r(\alpha)\right)\right]. \quad (3)$$
Note this replacement ensures that all probability values within the $\alpha$-cuts of $P(E|H)P(H)$ and $P(E)$ satisfy the required inequality (2) for every value of $\alpha$, while simultaneously minimizing the impact on the $\alpha$-cut intervals of $P(E)$.

A feature of this strategy, arising from the interval division formula, is that for every $\alpha$-cut such that the corresponding $\alpha$-cuts of $P(E|H)P(H)$ and $P(E)$ overlap, the right endpoint of the corresponding $\alpha$-cut of $P(H|E)$ attains unity value. In cases where a $P(E)$ $\alpha$-cut lies entirely within, or even to the left of the corresponding $\alpha$-cut of $P(E|H)P(H)$, the interval in (3) has zero width. The corresponding $\alpha$-cuts of $P(H|E)$ will then be given by:
$$\left[E_\ell(\alpha)/PE_r(\alpha), 1\right]. \quad (4)$$

Our approach is conservative in that it produces the lengthiest interval of feasible values for each $P(E)$ $\alpha$-cut that is consistent both with the original $P(E)$ $\alpha$-cuts and the inequality (2). This in turn maximizes the lengths of the corresponding $P(H|E)$ $\alpha$-cuts, to reflect a maximum degree of imprecision at each value of $\alpha$ where our adjustment in (3) is required. Thus, our results are both conservative and intuitive, as we illustrate in the examples below.

As a first example, assume that 6 SMEs provide the following interval estimates for $P(E|H)$, $P(H)$ and $P(E)$:

$$P(E|H): \begin{bmatrix} 0.4 & 0.6 \\ 0.55 & 0.65 \\ 0.5 & 0.6 \\ 0.45 & 0.65 \\ 0.5 & 0.65 \\ 0.45 & 0.65 \end{bmatrix} \quad P(H): \begin{bmatrix} 0.2 & 0.4 \\ 0.35 & 0.5 \\ 0.15 & 0.45 \\ 0.3 & 0.4 \\ 0.25 & 0.6 \\ 0.45 & 0.65 \end{bmatrix} \quad P(E): \begin{bmatrix} 0.4 & 0.6 \\ 0.4 & 0.5 \\ 0.4 & 0.7 \\ 0.4 & 0.8 \\ 0.35 & 0.75 \\ 0.45 & 0.65 \end{bmatrix}$$

Using the method described in Section III, Figure 1 shows the corresponding IT2 MFs for $P(E|H)$ (red FOU), $P(H)$ (purple FOU), $P(E)$ (blue FOU), and $P(E|H)P(H)$ (green FOU). Note that for $\alpha$ values up to about 0.09, the $P(E)$ UMF

overlaps the $P(E|H)P(H)$ UMF, which will produce the corresponding α-cuts in (4).

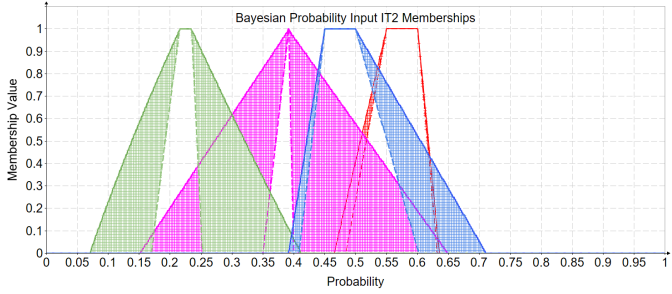

**Figure 1.** Input IT2 MFs for calculation of $P(H|E)$.

The resulting IT2 MF for $P(H|E)$ is shown in Fig. 2. The breadth of this FOU is dictated by the relatively broad input FOUs in Fig. 1. The two red diamonds at the top are the centroid interval endpoints as calculated from by the Enhanced Karnik-Mendel algorithm [29], while the blue diamond is the midpoint of the centroid interval, representing the best scalar reduction of this FOU. As noted above, the α-cuts for α values up to about 0.09 extend to unity as their righthand endpoints.

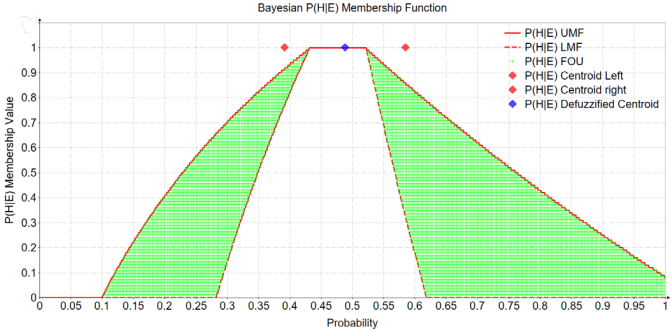

**Figure 2.** IT2 MF of $P(H|E)$ from Fig. 1 inputs.

As a second example, assume that the 6 SMEs provide the following interval estimates for $P(E|H)$, $P(H)$ and $P(E)$:

$$P(E|H): \begin{bmatrix} 0.4 & 0.5 \\ 0.55 & 0.6 \\ 0.5 & 0.55 \\ 0.45 & 0.55 \\ 0.5 & 0.6 \\ 0.45 & 0.5 \end{bmatrix} \quad P(H): \begin{bmatrix} 0.25 & 0.35 \\ 0.35 & 0.4 \\ 0.25 & 0.3 \\ 0.3 & 0.35 \\ 0.3 & 0.35 \\ 0.35 & 0.4 \end{bmatrix} \quad P(E): \begin{bmatrix} 0.55 & 0.6 \\ 0.45 & 0.5 \\ 0.45 & 0.55 \\ 0.5 & 0.6 \\ 0.45 & 0.55 \\ 0.55 & 0.65 \end{bmatrix}$$

Again using the method of Section III, Figure 3 shows the corresponding IT2 MFs for $P(E|H)$, $P(H)$, $P(E)$ and $P(E|H)P(H)$, using the same color scheme as in Fig. 1. Note that these IT2 MFs are narrower in their supports and the $P(E)$ MF does not overlap that of $P(E|H)P(H)$. Also note that all the FOUs have a solo peak value due to the *Null* overlap interval of their respective inputs.

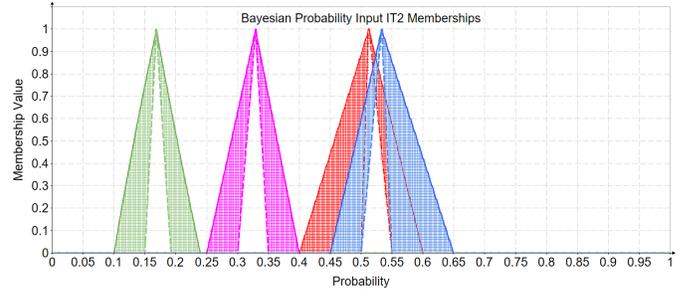

**Figure 3.** Input IT2 MFs for the second example.

The resulting IT2 MF for $P(H|E)$ is shown in Fig. 4. Note the considerably narrower width and solo peak for this FOU, due to the similar attributes of the inputs. Also, the centroid interval is narrower due to the generally slimmer tails of the input FOUs. Since the $P(E)$ FOU has no overlap with the $P(E|H)P(H)$ FOU, the resulting $P(H|E)$ can be computed from standard interval arithmetic operations upon the corresponding α-cuts, without needing the adjustments of (3).

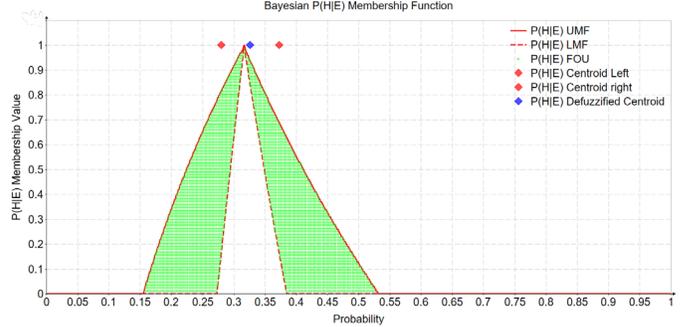

**Figure 4.** IT2 MF of $P(H|E)$ from Fig. 3 inputs.

These examples illustrate the utility of our generalization of Bayes' Theorem to accept arbitrary IT2 MFs for the input probabilities, even in those cases where the denominator MF overlaps with the numerator MF.

There remains the issue of how to map SME-provided interval ranges estimates of the requisite probabilities into IT2 MFs. In the next section, we present a novel algorithm for this task.

## III. SYNTHESIS OF SME INTERVAL INPUTS INTO INTERVAL TYPE-2 FUZZY MFs

The problem of encoding interval data received from one or more human subjects into IT2 fuzzy MFs has been addressed over the past two decades primarily in the context of computing with words (CWW) [21]-[28]. The basic procedure in all these instances is to collect interval data sets from multiple subjects who are members of the "general public", typically describing on a scale from 0 to 10 their interval range estimates corresponding to the words in a vocabulary.

For example, if the vocabulary describes size, the corresponding words might range from "teeny-weeny" to "tiny"

and up to "maximum amount" in a chosen vocabulary granulation, with the vocabulary granulation ranging typically from a minimum of three words up to a maximum of perhaps 32 words. Thus, the interval range estimates for the word "teeny-weeny" would generally correspond to small intervals lying near 0, while those of "maximum amount" would generally correspond to small intervals lying close to the upper bound of 10. In either of these two cases, the left (right) bound of the respective intervals will typically be 0 (10). For intermediate vocabulary words, e.g., "moderate amount", the intervals generally will lie in a middle range between 0 and 10.

The reason that interval sets for words are typically solicited from the general public is because words mean different things to different people, and even for the same person, they may have slightly different meanings over time. Thus, the inherent linguistic uncertainties captured in an IT2 MF are best derived from sampling a broad population of individuals.

However, this can present problems of "bad data" due to inputs from individuals who may either deliberately or unintentionally provide *prima facie* invalid intervals for a given word. For example, an interval having a left endpoint less than 0 or a right endpoint greater than 10 is impermissible for modeling on a scale of 0 to 10, while an interval encompassing large values near 10 would be clearly inappropriate to describe a word such as "teeny-weeny".

Thus, the approaches described in the above references typically precede their IT2 MF construction with a series of "data cleaning" steps to eliminate "bad" intervals from the interval sets corresponding to each word, with the remaining intervals after these cleaning operations being used to formulate IT2 MFs. Some of these steps are deterministic, while others involve statistical tests.

Of the above references on this topic, the approach described in [26] is perhaps considered the state of the art, since it produces "normal" IT2 MFs, i.e., having a unity membership value for both the upper and lower membership functions (UMF and LMF, respectively) over some subinterval for which all intervals in the cleaned interval set for a given word overlap.

The rationale for this construction feature is that, over any subinterval range for which all subjects are in agreement regarding the appropriate membership interval range for a word, there should be no imprecision in the constructed IT2 membership values (i.e., both the UMF and LMF values over this range should be unity, such that there is no primary or secondary uncertainty). Outside this subinterval of unity values, there generally will be a difference between the UMF and LMF values, with the area between these curves denoted as the "footprint of uncertainty" (FOU). Since the construction in [26] is considered the current state of the art, we shall adopt its feature of constructing normal IT2 MFs, but we focus our attention subsequently on a different construction method.

In [26], three types of trapezoidal IT2 MFs and their corresponding FOUs may result from the construction approach described therein. They are "Left shoulder", "Interior" and "Right shoulder" FOUs, respectively.

A Left shoulder FOU has unity values of both the UMF and LMF over some interval $[0,a]$ on the left end of the 0 to 10 scale, with UMF and LMF values linearly tapering to zero, such that for all $x > a$, the UMF value is generally greater than the LMF value so long as the UMF value exceeds zero. A Right shoulder FOU has unity values over some interval $[b,10]$ on the right end of the 0 to 10 scale, with UMF and LMF values linearly increasing from zero to unity at $b$, such that for all $x < b$, the UMF value is generally greater than the LMF value so long as the UMF value exceeds zero. An Interior FOU has a support interval for the UMF that is contained within the interval $[0,10]$ such that the UMF value is zero at both $x = 0$ and $x = 10$. Thus, an Interior FOU has UMF/LMF "tails" that taper to zero on each side of the interval for which the UMF and LMF have unity values, whereas a Left shoulder FOU has a single "tail" on its righthand side and a Right shoulder FOU has a single "tail" on its lefthand side.

## A. SME informed construction approach

The approach described in this paper differs from the one described above in three primary respects: 1) the input interval data sets used to formulate IT2 MFs come from SMEs rather than the general public, and thus we trust their inputs not to be frivolous or illegitimate; 2) we are not dealing with linguistic uncertainties and their associated imprecisions of meanings, but rather with physical and/or technical quantities (e.g., probabilities or odds) whose values are known only imprecisely, for which interval estimates can be solicited from SMEs; and 3) the intervals provided can be defined over arbitrary domains, which may be bounded or unbounded and positive or negative. Examples of the latter would be interval ranges describing probabilities, Moneyline odds, annual production quantities from a mining project, discount rates and appreciation factors for a commodity price, for which in some cases there are bounded ranges while in other cases no natural bounds can be specified.

This leads us to propose a novel method for the construction of IT2 MFs from sets of interval inputs such as those described above. Our method dispenses entirely with the "data cleaning" operations used in [26] and other word-based approaches, since we take the SME's inputs at face value.

We also generalize the class of FOUs that can be constructed to include MFs whose tails may intersect left or right boundaries of interval values (where they exist) at intermediate values of membership. We refer to these as "droop" FOUs, and thus in addition to Left/Right shoulder and Interior FOUs, we have "Left droop", "Right droop" and "Interior droop" FOUs, as described below. A key feature of the latter group of FOUs is that, unlike the first group mentioned previously, we are not required to set the boundary membership values at either 0 or 1 but can accommodate any values in the membership interval $[0,1]$.

Furthermore, our construction method inherently limits the outer boundaries of the UMF support interval to the most extreme interval boundary values present in a given input set of intervals provided by the SMEs, which is not the case with the word MF construction methods described in [26]. This is a more logical way of dealing with intervals describing physical or technical quantities based upon expert inputs, as we should respect the outer limits posed by the experts rather than allowing statistically based extensions of the UMF support interval. The latter might be acceptable when dealing with the imprecision associated with the meaning of words, but not when dealing with SME estimates of physical or technical quantities.

Finally, we provide a parametric means of adjusting the widths of the MF tails based upon the distribution of the interval boundaries, which allows us to determine the degree of secondary membership ranges in the MFs. The choice of parameter (as described below) can account for the level of secondary imprecision in an IT2 MF, i.e., the differences between the UMF and LMF for any value of the independent variable $x$. Typically, this choice would be based on a measure of consistency or inconsistency in the interval sets provided by the SMEs, as will described below.

### B. Construction method

Assume that, for a particular input quantity, we are provided $n$ SME interval inputs, where the $i^{th}$ SME provides an interval estimate $\left[a^{(i)},\, b^{(i)}\right]$ for that input quantity, $i = 1,\ldots,n$. Given this set of interval data, we perform the processing steps described in the following sequence.

*1) Determine if an overlap interval exists.*

To do this, we calculate $o_\ell = \max_i a^{(i)}$ and $o_r = \min_i b^{(i)}$. If $o_r > o_\ell$, then a nonzero length overlap interval $olap = \left[o_\ell,\, o_r\right]$ exists. Otherwise, $olap = Null$.

*2) Determine the FOU category.*

We consider two classes of FOUs: 1) ones having a natural minimum and maximum bound $\left[x_\ell,\, x_r\right]$, e.g., an interval of probability values, which must lie within $\left[0,\, 1\right]$; and 2) ones lacking one or more natural bounds, e.g., an interval of production forecasts for a mining property.

To assign a set of intervals to an FOU category, the first task is to determine if natural bounds to the minimum/maximum values exist and, if so, determine if the left/right bounds of one or more of the intervals have left endpoints equal to the lower bound $x_\ell$ or if one or more of the intervals have right endpoints equal to the upper bound $x_r$. We do this by performing the calculations shown in Algorithm 1, where only the calculation for $x_\ell$ or for $x_r$ is performed for intervals having only one natural bound. Note that the resulting values $x_0$ and $x_1$ represent the fractions of intervals whose left (right) endpoints intersect the vertical axes $x = x_\ell$ and $x = x_r$, respectively.

An Interior droop FOU is assigned if $0 < x_0 < 1$ and $0 < x_1 < 1$, i.e., some but not all intervals intercept both the $x = x_\ell$ and $x = x_r$ vertical axes. A Left droop FOU is assigned in cases where $0 < x_0 < 1$ but $x_1 = 0$, i.e., some but not all intervals intersect the $x = x_r$ vertical axis, but none intersect the $x = x_r$ vertical axis when this bound exists. (This type of FOU generalizes the Left shoulder FOU by allowing for a "drooping" shoulder.) Similarly, a Right droop FOU is assigned in cases where $x_0 = 0$ but $0 < x_1 < 1$, i.e., some but not all intervals intersect the $x = x_r$ vertical axis, but none intersect the $x = x_\ell$ vertical axis when this bound exists.

Assign a Left shoulder FOU if the overlap interval is non-$Null$ and the left endpoint of the overlap interval $o_\ell = x_\ell$, i.e., *all* intervals have a left endpoint of $x_\ell$. Similarly, assign a Right shoulder FOU if the right endpoint of the overlap interval $o_r = x_r$, i.e., *all* intervals have a right endpoint of $x_r$. In all remaining cases other than those described above, assign an Interior FOU.

---

**Algorithm 1: Calculating Vertical Axis Intercepts**

**Input**: Left and/or right domain bounds $x_\ell$, $x_r$

SME input intervals $\left[a^{(i)},\, b^{(i)}\right], i = 1,\ldots,n$

**Output**: The fraction $x_0$ of SME input intervals having left endpoints equal to $x_\ell$ and/or the fraction of SME input intervals having right endpoints equal to $x_r$

**Initialize**: $x_0 = 0;\ x_1 = 0$

**for** $i = 1,\ldots,n$

    **if** $x_\ell > -\infty \wedge a^{(i)} = x_\ell$

        $x_0 = x_0 + 1$

    **if** $x_r < \infty \wedge b^{(i)} = x_r$

        $x_1 = x_1 + 1$

**end**

$x_0 = \dfrac{x_0}{n};\ x_1 = \dfrac{x_1}{n}$

**Return** $x_0,\ x_1$

---

*3) Create reduced interval sets.*

For interval sets having a non-$Null$ overlap, this is done by removing the overlap interval from each interval, which results in a single set of reduced intervals for Left or Right shoulder FOUs. For all other FOU categories, it results in two sets of reduced intervals. Thus, for Left shoulder FOUs, we have the reduced interval set as $\left[o_r,\, b^{(i)}\right]$ for $i = 1,\ldots,n$. For Right shoulder FOUs, the reduced interval set is $\left[a^{(i)},\, o_\ell\right]$ for $i = 1,\ldots,n$. For all other cases of non-$Null$ overlap, we have two reduced interval sets, one set lying to the left of the overlap interval, i.e., $\left[a^{(i)}, o_\ell\right]$ for $i = 1,\ldots,n$, and one set lying to the right of the overlap interval, i.e., $\left[o_r, b^{(i)}\right]$ for $i = 1,\ldots,n$.

In cases where the overlap is $Null$, we first find the mean $m$ of all interval endpoints (which is identical to the mean of the interval centers). For any interval lying to the left (inclusive) of the mean, we assign that interval to the reduced left interval set, while for any interval lying to the right (inclusive) of the mean, we assign that interval to the reduced right interval set. For intervals that straddle the mean, we split them into two intervals, with the left portion of the interval assigned to the reduced left interval set, and the right portion of the interval assigned to the reduced right interval set. This results in a reduced left interval set $\left[a^{(j)}, c^{(j)}\right]$ for $j = 1,\ldots,n_\ell$ where $c^{(j)} \leq m$ for all $j$, and a reduced right interval set $\left[d^{(k)}, b^{(k)}\right]$ for $k = 1,\ldots,n_r$, where $d^{(k)} \geq m$ for all $k$. Note that $c^{(j)} = b^{(j)}$ for intervals lying wholly to the left of $m$ and $d^{(k)} = a^{(k)}$ for intervals lying wholly to the right of $m$, i.e., these intervals are not split. For intervals that are split, $c^{(i)} = d^{(i)} = m$.

In some cases, we may have intervals that lie entirely in the negative half plane and/or that overlap and/or lie on both sides of the $x = 0$ axis. In these cases, we first segment the intervals into two groups, with one group comprising all intervals or segments of intervals having nonnegative ranges and the other group comprising all intervals or segments of intervals having nonpositive ranges. Thus, intervals that straddle $x = 0$ are split into two intervals with a common endpoint of $x = 0$. We then apply the approach described below to each group, where for the nonpositive intervals we first mirror these intervals about the $x = 0$ axis before performing our calculations and then invert the results back to the negative half-axis.

Calculate the trapezoidal parameters $\ell b$, $\ell t$, $rt$ and $rb$ (for left-bottom, left-top, right-top and right-bottom, respectively) for a truncated trapezoid function $trap(x, h, \ell b, \ell t, rt, rb, x_0, x_1)$ of height unity (i.e., $h = 1$), where $trap(x, h, \ell b, \ell t, rt, rb, x0, x1)$ is defined by:

$$trap(x, h, \ell b, \ell t, rt, rb, x_0, x_1) = \begin{cases} 0, & x < x_0 \\ \dfrac{h}{(\ell t - \ell b)} x - \dfrac{\ell b * h}{(\ell t - \ell b)}, & x_0 \leq x \leq \ell t \\ 1 & \ell t < x < rt \\ \dfrac{-h}{(rb - rt)} x + \dfrac{rb * h}{(rb - rt)}, & rt \leq x \leq x_1 \\ 0 & x > x_1 \end{cases} \quad (5)$$

For reduced interval sets where no interval intersects the left or right bounds $x_0$ or $x_1$, the trapezoidal parameters for the UMF and LMF are calculated as the weighted power mean (WPM) $wpm(x, w, r)$ of the interval endpoints, using unity weights, i.e., $w_i = 1$ for all $i$. The WPM is specified by (where $x$ and $w$ are vectors with elements $x_i$ and $w_i$):

$$wpm(x, w, r) = \lim_{p \to r} \left[\frac{1}{\sum_i w_i} \left(\sum_i w_i x_i^p\right)^{\frac{1}{p}}\right], \quad (6)$$

which is directly evaluated numerically for values of $x_i \in [0,1]$.

However, since we may encounter arbitrarily large values of $x_i$ for quantities such as production profiles, to avoid numerical overflow problems we first normalize the $x_i$ values in (6) by dividing them by $x_{\max} = \max_i x_i$, multiplying the sum by $x_{\max}^p$, then calculating the natural log of the resulting expression within brackets (which is numerically stable) and finally taking the exponential of this result. With some further treatment of limiting cases, we arrive at the following revised formula, which is equivalent to (6) but is numerically well-behaved for large $x$ values:

$$wpm(x, w, r) = \begin{cases} 0, & x_i = 0 \wedge r \leq 0 \\ x_{\max} \exp\left[\dfrac{1}{r} \ln\left[\dfrac{1}{\sum_i w_i} \sum_i w_i \left(\dfrac{x_i}{x_{\max}}\right)^r\right]\right], & 0 < |r| < \infty \\ \prod_i (x_i)^{\frac{w_i}{\sum w_i}}, & r = 0 \\ \min_i x_i, & r = -\infty \\ \max_i x_i, & r = \infty \end{cases} \quad (7)$$

The WPM exhibits several useful properties for particular values of $r$ when all the weights $w_i = 1$. From the last two lines of (7), we see that for the $\pm\infty$ extreme values of $r$, it yields the minimum (for $r = -\infty$) or maximum (for $r = \infty$) of the $x_i$. For $r = 1$, it yields the arithmetic average of the $x_i$. For $r = 2$, it yields the root mean square value of the $x_i$. For $r = 0$, it yields the geometric mean of the $x_i$. For $r = -1$, it yields the harmonic mean of the $x_i$. Furthermore, $wpm(x, w, r)$ increases monotonically with $r$ when all $x_i > 0$. Thus, the WPM is a very useful aggregation function for nonnegative values of $x_i$, which suggests its use in the aggregation of interval endpoints when an FOU tail lies completely within the bounding interval $[x_\ell, x_r]$.

To apply the WPM in the case of a Left shoulder FOU, we calculate the trapezoidal parameters (see (5)) of the UMF as follows (where $w$ is the unity vector):

$$\ell b = x_0, \quad (8)$$

$$\ell t = x_0, \quad (9)$$

$$rt = o_r \quad (10)$$

$$rb = wpm(b^{(i)}, w, r), \ r \geq 1. \quad (11)$$

For a Left shoulder LMF, we calculate:

$$\ell b = x_0, \quad (12)$$

$$\ell t = x_0, \quad (13)$$

$$rt = o_r \quad (14)$$

$$rb = wpm(b^{(i)}, w, 2-r), \ r \geq 1. \quad (15)$$

Given the monotonicity of the WPM with $r$, we are assured that the UMF values will always be greater than or equal to the LMF values. For $r = 1$, the $rb$ values in (11) and (15) are equal, which collapses the FOU to a type-1 fuzzy MF. For $r = \infty$, the $rb$ value for the UMF is the maximum of the righthand interval endpoints, while the $rb$ value for the LMF is the minimum of these interval endpoints, which results in an FOU of maximal tail width.

Thus, by choosing the value of $r > 1$, we control the tail width of the FOU as a function of the distribution of the right endpoints of the reduced interval set. The broader the latter distribution and the larger the value of $r$, the "fatter" the tail. Conversely, if the distribution of interval endpoints is narrow and/or $r$ is close to unity, the tail width will be narrow.

Note that, in the extreme case where all right endpoints of the reduced intervals are identical, all values of the WPM are identical for all $r$, and thus the $rb$ values in (11) and (15) are equal, and in fact we have $rb = rt = o_r$ resulting in a unity-valued MF of the FOU for the overlap interval. This makes intuitive sense, since in this case all SMEs are in complete agreement regarding their interval endpoint estimates, and there is a uniform, unity-valued membership for all values of $x$ in the interval.

For a Right shoulder FOU, we calculate the trapezoidal parameters of the UMF as follows, where again $w$ is the unity vector:

$$\ell b = wpm(a^{(i)}, w, 2-r), \ r \geq 1, \quad (16)$$

$$\ell t = o_\ell, \quad (17)$$

$$rt = x_1, \quad (18)$$

$$rb = x_1. \quad (19)$$

For the Right shoulder LMF, we have:

$$\ell b = wpm(a^{(i)}, w, r), \ r \geq 1, \quad (20)$$

$$\ell t = o_\ell, \quad (21)$$

$$rt = o_r, \quad (22)$$

$$rb = o_r. \quad (23)$$

Note that for the $\ell b$ values in this case, we reverse the WPM exponents for the UMF and LMF, so that the $\ell b$ value for the LMF is greater than that of the UMF, as would be expected.

For Interior FOUs, there will be both a lefthand and righthand tail, with the trapezoid parameters for $\ell b$ determined by (16) and (20), while those for $rb$ are determined by (11) and (15). The $\ell t$ and $rt$ values are of course just the overlap interval endpoints $o_\ell$ and $o_r$, respectively.

For droop FOUs, the left and/or right tail of the UMF intersects the $x = x_\ell$ and/or $x = x_r$ vertical axes at some intermediate value(s) between 0 and 1. Figure 5 illustrates this situation for an Interior droop FOU, which has droop tails on each side. Since this is the more general case (Left/Right droop FOUs have just one such tail on their right/left side, respectively), we address it first.

Our construction method is to set the intercept of the left (right) tail of the UMF with the $x = x_\ell$ $(x = x_r)$ vertical axis equal to the fraction of intervals in the reduced interval sets for which $x_\ell$ $(x_r)$ is their left (right) endpoint, respectively, as determined by Algorithm 1. Thus, in Figure 5 we designate the left/right $x$-intercept values of the UMF tails by $\overline{x}_\ell$ and $\overline{x}_r$, respectively. The $\ell t$ and $rt$ values correspond to the overlap interval endpoints, i.e.,

$$\ell t = o_\ell, \quad (24)$$

$$rt = o_r, \quad (25)$$

where in the case where the overlap is *Null*, we have $\ell t = rt = m$.

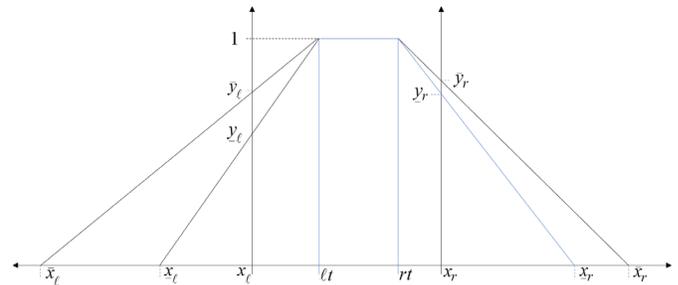

**Figure 5**. Construction method for droop FOUs

Thus, we have two points lying on the linear UMF tails on the left and right sides, from which we can determine the equations of each UMF tail. There remains the question of what should be the intercept values $\underline{y}_\ell$ and $\underline{y}_r$ of the LMF with the $x = x_\ell$ and/or $x = x_r$ vertical axes. In keeping with our use of the WPM to provide a flexible aggregation operator, we propose the following values:

$$\underline{y}_\ell = wpm\left(\begin{bmatrix}\overline{y}_\ell \\ 0\end{bmatrix}, \begin{bmatrix}1 \\ 1\end{bmatrix}, r0\right), \quad (26)$$

$$\underline{y}_r = wpm\left(\begin{bmatrix}\overline{y}_r \\ 0\end{bmatrix}, \begin{bmatrix}1 \\ 1\end{bmatrix}, r1\right). \quad (27)$$

In other words, we use the WPM to determine a value between 0 and the UMF intercept with the vertical axis on the left (right) side, where the choice for the value of $r0$ ($r1$) over the range from $-\infty$ to $\infty$ determines how close the left (right) LMF intercept is to 0 or to $\overline{y}_\ell$ ($\overline{y}_r$). We suggest using values of $r0 = r1 = 1$, corresponding to the arithmetic average between 0 and $\overline{y}_\ell$ ($\overline{y}_r$), but this is entirely at the designer's choice. With these preliminaries, we can compute the remaining trapezoidal parameters as follows.

From Figure 5, the slope of the left side of the UMF is:
$$\overline{m}_\ell = \frac{1 - \overline{y}_\ell}{\ell t - x_\ell}, \tag{28}$$
while that of the right side of the UMF is:
$$\overline{m}_r = \frac{\overline{y}_r - 1}{x_r - rt}. \tag{29}$$
Similarly, the slope of the left side of the LMF is:
$$\underline{m}_\ell = \frac{1 - \underline{y}_\ell}{\ell t - \underline{x}_\ell}, \tag{30}$$
while that of the right side of the LMF is:
$$\underline{m}_r = \frac{\underline{y}_r - 1}{\underline{x}_r - rt}. \tag{31}$$

The equation of the left side of the UMF is thus:
$$y = \overline{m}_\ell x + \overline{b}_\ell, \text{ where } \overline{b}_\ell = \overline{y}_\ell - \overline{m}_\ell x_\ell, \tag{32}$$
and the equation of the left side of the LMF is:
$$y = \underline{m}_\ell x + \underline{b}_\ell, \text{ where } \underline{b}_\ell = \underline{y}_\ell - \underline{m}_\ell \underline{x}_\ell. \tag{33}$$
The equation of the right side of the UMF is:
$$y = \overline{m}_r x + \overline{b}_r, \text{ where } \overline{b}_r = \overline{y}_r - \overline{m}_r x_r, \tag{34}$$
while that of the right side of the LMF is:
$$y = \underline{m}_r x + \underline{b}_r, \text{ where } \underline{b}_r = \underline{y}_r - \underline{m}_r \underline{x}_r. \tag{35}$$
From (32), we have when $y = 0$,
$$\overline{x}_\ell = -\frac{\overline{b}_\ell}{\overline{m}_\ell}, \tag{36}$$
and from (33) when $y = 0$,
$$\underline{x}_\ell = -\frac{\underline{b}_\ell}{\underline{m}_\ell}. \tag{37}$$
From (34), we have when $y = 0$,
$$\overline{x}_r = -\frac{\overline{b}_r}{\overline{m}_r}, \tag{38}$$
and from (35), when $y = 0$, we have
$$\underline{x}_r = -\frac{\underline{b}_r}{\underline{m}_r}. \tag{39}$$

The cases of Left (Right) droop FOUs are handled in analogous fashion for their left (right) tail.

In the case where all intervals lie in the nonpositive half-plane, we mirror the interval values about $x = 0$, calculate the trapezoidal parameters on these nonnegative values, then take the negative of these values and reverse the order of the trapezoid parameters. In cases where intervals lie in both half-planes (including cases where some intervals straddle $x = 0$), we perform the calculations separately on the nonnegative and nonpositive intervals, then append the right trapezoid tail of the nonnegative group to the right overlap interval endpoint or the mean for *Null* overlap and append the (reversed) right trapezoid tail of the mirrored nonpositive group to the left overlap interval endpoint (or the mean for *Null* overlap).

Thus, with the above construction, we can derive the resulting FOUs from any set of intervals provided by SMEs.

We mention in passing the exceptional case where only a single SME is available to provide information. In that case, we follow [26] and request that they provide two interval range estimates of the minimum and maximum value for a parameter of interest. We then generate multiple uniform random pairs drawn from these intervals and use those pairs as interval endpoints in the above method.

As a final issue, there remains the appropriate choice of the WPM exponent $r \geq 1$ to use in the above methods. One suggested approach is that the value of $r$ be determined by the following measure of interval consistency:
$$r = \begin{cases} \dfrac{b_{\max} - a_{\min}}{o_r - o_\ell}, & olap \neq Null \\ \infty, & olap = Null \end{cases} \tag{40}$$
where
$$b_{\max} = \max_i b^{(i)}, \tag{41}$$
$$a_{\min} = \min_i a^{(i)}, \tag{42}$$
and $o_\ell$ and $o_r$ are the respective left- and right-hand bounds of the overlap interval $olap$. Thus, in the (unlikely) event that all intervals in a set are identical, we obtain $r = 1$ (which results in a type-1 MF with no tail width), while if the overlap interval is *Null* (i.e., no common agreement between SMEs), then $r = \infty$, resulting in maximum tail width. Between these two extremes, the value of $r$ increases from unity as the ratio of the width of the maximum range of the intervals to the width of the overlap interval. Thus, the less consistent the intervals, the greater the value of $r$ and the fatter the tail(s) of the FOU.

## IV.    EXAMPLES OF CONSTRUCTION APPROACH

In this section, we present some examples of IT2 MFs constructed using the methods of the previous section. In these examples, we select an arbitrary choice of $r = 10$ for the value of the WPM exponent for purposes of illustration, whereas in practice, we might use the value suggested by (40) above.

We begin with an example of SME-provided intervals corresponding to a production profile for a particular year. Suppose that five SMEs provide the following set of intervals for the units of production of a commodity in a given year:

$$pr = \begin{bmatrix} 900,000 & 1,200,000 \\ 850,000 & 1,100,000 \\ 850,000 & 1,300,000 \\ 1,000,000 & 1,500,000 \\ 800,000 & 1,300,000 \end{bmatrix}. \quad (43)$$

The overlap interval for this data is $[1,000,000,\ 1,100,000]$, and since none of the interval endpoints intersect any natural boundaries, the category of the corresponding FOU is an Interior FOU. Upon removing the overlap interval, the two sets of reduced intervals to the right and left of the overlap interval are:

$$pr_\ell = \begin{bmatrix} 900,000 & 1,000,000 \\ 850,000 & 1,000,000 \\ 850,000 & 1,000,000 \\ 1,000,000 & 1,000,000 \\ 800,000 & 1,000,000 \end{bmatrix} \quad pr_r = \begin{bmatrix} 1,100,000 & 1,200,000 \\ 1,100,000 & 1,100,000 \\ 1,100,000 & 1,300,000 \\ 1,100,000 & 1,500,000 \\ 1,100,000 & 1,300,000 \end{bmatrix}. \quad (44)$$

If we choose $r = 10$ for the WPM exponent in (11), (15), (16) and (20) to compute the values of $\ell b$ and $rb$ for the UMF and LMF of the Interior FOU, we obtain the IT2 MF shown in Figure 6.

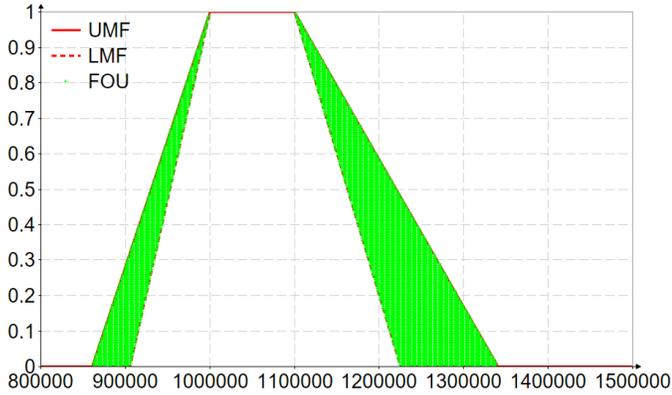

**Figure 6.** IT2 MF for intervals in (43), $r = 10$.

As $r \to 1$ from above, the UMF and LMF tails in Figure 6 will converge to a single line corresponding to a type-1 MF, which represents one extreme for the choice of $r$, with a corresponding zero tail "fatness" of the FOU, i.e., a type-1 MF. In this case, the values of $\ell b$ and $rb$ for both the UMF and LMF are the arithmetic average of the left and right interval endpoints, respectively.

At the opposite extreme, as $r \to \infty$, the UMF support in stretches to the full x-axis range of $[800,000,\ 1,500,000]$ (i.e., the min and max interval endpoints in (43)), while the LMF support collapses to a rectangular function whose support interval is simply the overlap interval. The latter FOU would represent the maximum tail fatness obtainable with our method and corresponds to the largest secondary imprecision of the membership values. For intermediate values of $1 \leq r < \infty$, the tail fatness varies continuously as a function of $r$. This illustrates the flexibility of our method in the selection of the value of $r$ to sculpt the secondary imprecision (i.e., tail fatness) of the IT2 MF.

Now suppose that we modify the production profile interval set in (43) to have the following interval values, which have a *Null* overlap:

$$pr1 = \begin{bmatrix} 900,000 & 1,200,000 \\ 850,000 & 1,100,000 \\ 850,000 & 1,200,000 \\ 1,400,000 & 1,600,000 \\ 800,000 & 1,300,000 \end{bmatrix}. \quad (45)$$

The two sets of reduced intervals to the left and right of the mean value $m = 1,120,000$ of the interval endpoints in (45) are:

$$pr1_\ell = \begin{bmatrix} 900,000 & 1,120,000 \\ 850,000 & 1,100,000 \\ 850,000 & 1,120,000 \\ 800,000 & 1,120,000 \end{bmatrix} \quad pr_r = \begin{bmatrix} 1,120,000 & 1,200,000 \\ 1,120,000 & 1,200,000 \\ 1,400,000 & 1,600,000 \\ 1,120,000 & 1,300,000 \end{bmatrix}. \quad (46)$$

Note that the reduced interval sets each have four intervals, since the second interval in (45) lies entirely to the left of $m$ and the fourth interval in (45) lies entirely to the right of $m$. Thus, these two intervals are not split into two intervals because they do not straddle the mean $m$, and therefore they reduce the interval count of the reduced interval sets. The corresponding FOU remains an Interior FOU, since the interval endpoints do not intersect any natural boundaries, but it is now a triangular shaped FOU with a peak at the mean $m$.

For the choice of $r = 10$, we obtain the FOU shown in Figure 7. The narrowness of the left tail of this FOU is due to the relatively narrow range of the left-hand reduced interval set left endpoints, which only spans a range of 100,000 (i.e., 800,000 to 900,000). In contrast, the wider righthand tail is due to the larger range of the righthand reduced interval set right interval endpoints, which spans a range of 400,000 (i.e., 1,200,000 to 1,600,000). This illustrates how the range of interval endpoints affects the tail fatness for a given value of $r$.

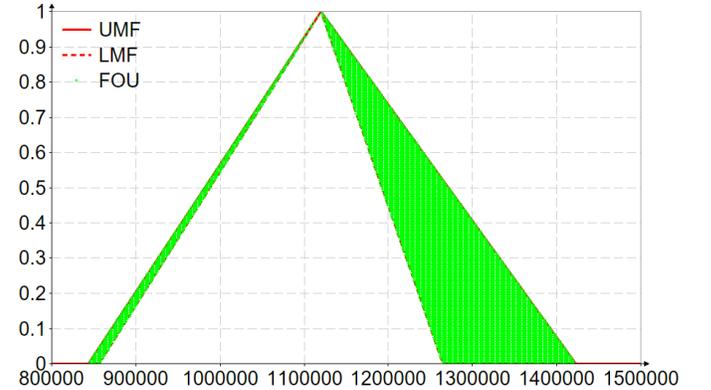

**Figure 7.** IT2 MF for intervals in (45), using $r = 10$.

We turn now to the case where there are both left and right natural interval bounds and where one or more interval endpoints intersect these bounds. This generally results in a "droop" type FOU, with the Left/Right shoulder FOUs being special cases of the former. For the Left/Right droop FOU cases, consider the following interval sets, where the natural bounds are $x_0 = 0$ and $x = 1$ (e.g., sets of probability intervals):

$$leftdroop = \begin{bmatrix} 0 & 0.3 \\ 0 & 0.5 \\ 0 & 0.3 \\ 0.1 & 0.5 \\ 0.2 & 0.4 \end{bmatrix} \quad rightdroop = \begin{bmatrix} 0.4 & 0.7 \\ 0.6 & 1 \\ 0.8 & 1 \\ 0.5 & 1 \\ 0.7 & 0.9 \end{bmatrix} \quad (47)$$

For the $leftdroop$ interval set, we observe that there is an overlap interval of $[0.2, 0.3]$, while for the $rightdroop$ interval set, the overlap is $Null$. Figure 8 shows the Left droop FOU for $r = 10$ resulting from the $leftdroop$ interval set, and Figure 9 shows the Right droop FOU for the same $r = 10$ resulting from the $rightdroop$ interval set in (47), where in both cases, the WPM exponent for the $y$-axis intercepts are given by $r0 = r1 = 1$.

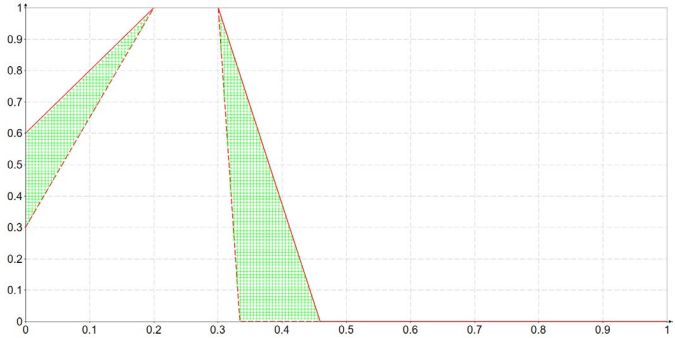

**Figure 8**. Left droop IT2 MF for intervals in (47), $r = 10$.

For the Left droop FOUs in Fig. 8, we observe that the $y$-axis intercept at $x = 0$ of the UMF occurs at a value of 0.6, which corresponds to the fraction of the $leftdroop$ intervals with endpoints equal to 0. Since there is a non-$Null$ overlap, the lefthand and righthand reduced interval sets both contain five intervals, with three of the intervals in the lefthand reduced interval set having a left endpoint of 0. For the $rightdroop$ interval set of Fig. 9, which has a $Null$ overlap, the reduced interval sets will each contain four intervals, with three of the four righthand intervals having a right interval endpoint of 1. Thus, the $y$-axis intercept of the UMF at $x = 1$ is 0.75. Since $r0 = r1 = 1$, the $y$-axis intercept of LMF is one-half the value of that of the UMF, since this choice of WPM intercept exponents corresponds to the arithmetic average of the UMF intercept and 0.

Further, we consider an example of an Interior droop interval data set, as follows:

$$indroop = \begin{bmatrix} 0 & 0.3 \\ 0 & 0.5 \\ 0 & 0.6 \\ 0.5 & 1 \\ 0.7 & 0.9 \end{bmatrix}. \quad (48)$$

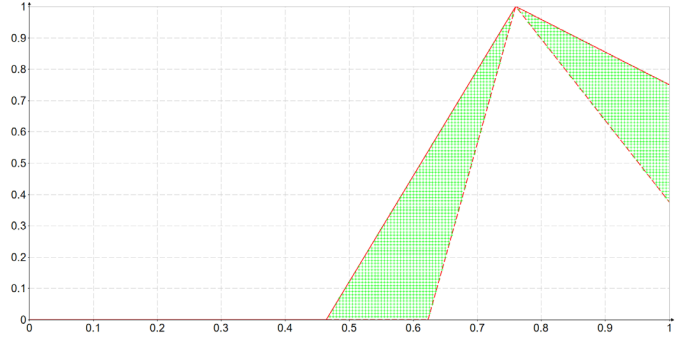

**Figure 9**. Right droop IT2 MF for intervals in (47), $r = 10$.

Note that the overlap in this set is $Null$ and the mean $m$ of all interval endpoints is 0.45. The second and third intervals straddle this mean value, so these intervals are split, resulting in a lefthand interval set having three intervals, all of which have a zero left endpoint, and a righthand interval set having four intervals, one of them with a right endpoint of unity value.

The resulting FOU is shown in Figure 10. Since all three intervals in the lefthand reduced interval set have left endpoints on the $y$-axis at $x = 0$, the UMF will have a unity value at $x = 0$. And since we continue to use $r0 = r1 = 1$ for the WPM exponents of the intercepts, the LMF has a value of 0.5 at $x = 0$. For the righthand reduced interval set, one of the four intervals in this set has a right endpoint of 1, so the UMF value at $x = 1$ is 0.25, while the LMF value at $x = 1$ is one-half of this value, or 0.125.

We can broaden or narrow the left or right tail widths by the selection of larger or smaller values of $r0$ or $r1$, respectively. As either exponent approaches $+\infty$, the corresponding LMF will converge to the UMF, resulting in a type-1 MF, while as either exponent approaches $-\infty$, the corresponding LMF $y$-axis intercept at $x = 0$ or $x = 1$ will approach 0, resulting in the fattest tail.

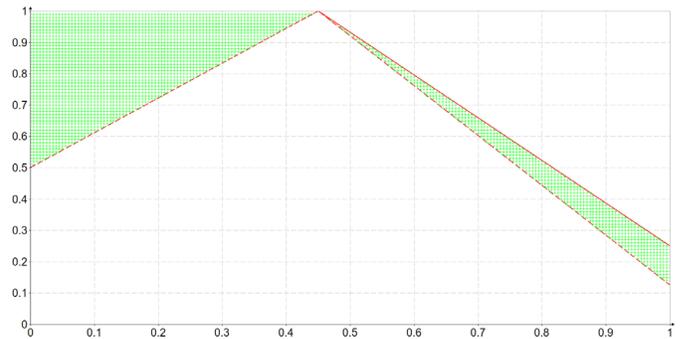

**Figure 10**. Interior droop IT2 MF for intervals in (48), $r = 10$.

As final examples, suppose that we have sets of Moneyline odds on a game, where for the projected winner the negative opening line and closing line odds might take interval ranges derived from 10 different sports books:

$$\text{opening} = \begin{bmatrix} -430 & -389 \\ -394 & -356 \\ -383 & -347 \\ -394 & -356 \\ -451 & -408 \\ -441 & -399 \\ -441 & -399 \\ -420 & -380 \\ -420 & -380 \\ -478 & -432 \end{bmatrix} \quad \text{closing} = \begin{bmatrix} -357 & -323 \\ -394 & -356 \\ -352 & -318 \\ -499 & -451 \\ -357 & -323 \\ -420 & -380 \\ -420 & -380 \\ -367 & -332 \\ -420 & -380 \\ -407 & -369 \end{bmatrix} \quad (49)$$

The resulting FOUs of the aggregate odds are shown in Fig. 11, with the opening line odds in green and the closing line odds in orange. This case illustrates the application of our approach to negative intervals.

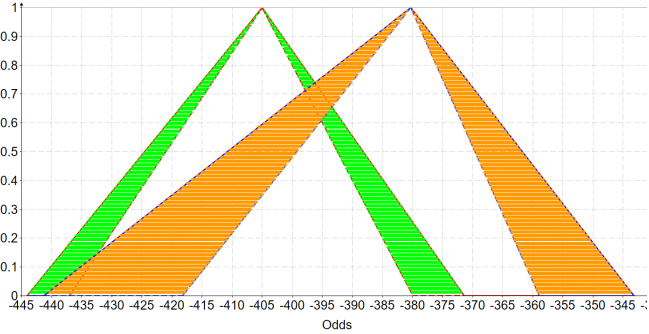

Figure 11. IT2 MFs for intervals in (49), $r = 10$.

Finally, we consider a case where we have a combination of positive and negative odds intervals:

$$\text{opening} = \begin{bmatrix} 95 & 105 \\ 95 & 105 \\ 95 & 105 \\ 101 & 111 \\ -110 & -100 \\ 102 & 112 \\ -115 & -104 \\ 102 & 112 \\ -110 & -100 \\ 106 & 118 \end{bmatrix} \quad \text{closing} = \begin{bmatrix} 104 & 115 \\ 100 & 110 \\ 104 & 115 \\ 101 & 111 \\ 99 & 109 \\ 95 & 105 \\ 95 & 105 \\ -111 & -199 \\ 95 & 105 \\ -106 & -96 \end{bmatrix} \quad (50)$$

Here, three of the 10 oddsmakers narrowly project the team to win (i.e., negative intervals) on their opening line, while the other 7 project it to lose (i.e., positive intervals). However, on the closing line, two different oddsmakers project a win while the remaining ones project a loss. Also, note that the odds intervals are quite close to one another, with relatively little variation in the endpoints for both the positive and negative intervals, respectively. The resulting FOUs are shown in Fig. 12, again with the green FOU corresponding to the opening line and the orange FOU corresponding to the closing line. Each FOU exhibits a wide range due to the presence of both positive and negative intervals that contribute to it. Also, each FOU has very narrow tails due to the closeness of the input odds intervals, which results in nearly type-1 MFs.

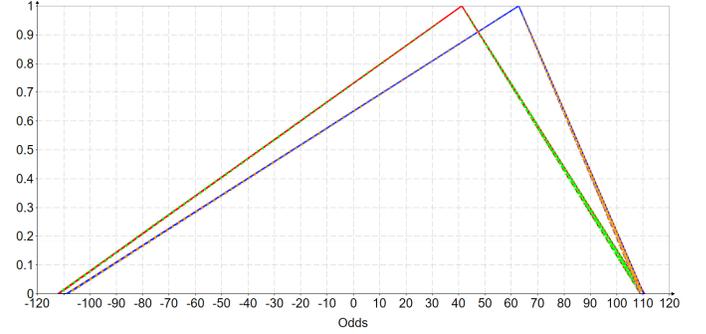

Figure 12. IT2 MFs for intervals in (50), $r = 10$.

## V. CONCLUSION

This work makes two key contributions. First, it provides a generalized version of Bayes' Theorem using IT2 MFs for the input probabilities $P(E|H)$, $P(H)$ and $P(E)$, while producing an IT2 MF of the output probability $P(H|E)$. These input probability IT2 MFs can be specified directly or derived from SME-provided interval range estimates. We also provide a conservative remedy for the problem of overlap between the resulting $P(E|H)P(H)$ and $P(E)$ FOUs, remedying the invalid probability values that would result from this overlap.

The resulting FOU for $P(H|E)$ is an information-rich display of the primary and secondary imprecision of this probability, which can be type-reduced to an interval and/or scalar value for simpler representations. Since real-world input probabilities to Bayes' Theorem are never known precisely, our approach enables the use of the best available information from SMEs. Our results may also provide some reconciliation of the historical enmity between Bayesian and fuzzy system theorists [30] via direct intermarriage between the two technologies.

Our second contribution is to introduce a novel algorithm for encoding interval inputs from SMEs into corresponding IT2 MFs. Our algorithm generalizes previous approaches that were aimed toward encoding word MFs from interval range estimates in $[0,10]$ provided by the general public. It allows for unbounded input interval ranges, so that IT2 MFs for any physical or technical quantity can be encoded. It further introduces three "droop" classes of MFs for cases where some fractions of the input intervals have left or right endpoints equal to their corresponding boundary values.

Given the ubiquitous employment of BT in many different domains and applications, we believe these results will be quite useful. In future work, we will apply our approach to specific problem domains where BT is used.

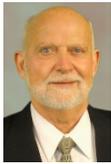
**John T. (Terry) Rickard** (S'67-M'75-SM'01) received the B.S. and M.S. degrees in electrical engineering from Florida Institute of Technology, Melbourne, Florida, in 1969 and 1971, respectively, and a Ph.D. degree in engineering physics from the University of California at San Diego, La Jolla, California, in 1975. In 1975, he co-founded ORINCON Corporation, a company specializing in the design and development of state-of-the-art data and information processing solutions for government and commercial customers. From 1994 to 2001, he served as President and later Chief Scientific Officer of OptiMark Technologies, Inc. In 2003, he became a Senior Principal Research Scientist at Lockheed Martin and was elected a Senior Fellow in 2005. He is now the Chief Data Scientist for Meraglim Holdings Corporation and a Managing Partner in Royalty & Streaming Advisors LLC. His current research interests are in computational intelligence, particularly type-2 fuzzy sets.

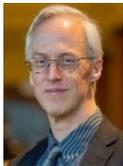
**William A. Dembski** (M'06-SM'21) received the B.A. degree in psychology from the University of Illinois at Chicago in 1981 and the S.M. and Ph.D. degrees in mathematics from the University of Chicago in 1985 and 1988, respectively. He also holds advanced degrees in philosophy and theology. The author and/or editor of more than 25 books, he has published in the peer-reviewed mathematics, engineering, and biology literature, as well as in the humanities literature. He founded and headed the first intelligent design think tank at a research university, Baylor's Michael Polanyi Center (1999-2000). He has been an NSF doctoral and post-doctoral fellow. Along with Francis Crick, Charles Townes, and Steven Weinberg, he is a winner of Texas A&M's Trotter Prize (awarded in 2005). He is a founding senior fellow of Discovery Institute's Center for Science and Culture. His main research these days focuses on extending his work on design-inferential reasoning. Besides his straight scientific and academic research, he is also an entrepreneur who builds educational websites and technologies.

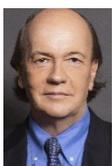
**James Rickards** (M, '23) received an LL.M. (Taxation) from the NYU School of Law; a J.D. from the University of Pennsylvania Law School; an M.A. in international economics from SAIS, and a B.A. (with honors) from Johns Hopkins. He is the Editor of *Strategic Intelligence,* a financial newsletter. He is the New York Times bestselling author of *MoneyGPT* (2024), *Sold Out* (2022), *The New Great Depression* (2021), *Aftermath* (2019), *The Road to Ruin* (2016), *The New Case for Gold* (2016), *The Death of Money* (2014), and *Currency Wars* (2011). He is an investment advisor, lawyer, inventor, and economist, and has held senior positions at Citibank, Long-Term Capital Management, and Caxton Associates. In 1998, he was the principal negotiator of the rescue of LTCM sponsored by the Federal Reserve. He is an op-ed contributor to the *Financial Times*, *Evening Standard*, *The Telegraph*, *New York Times*, and *Washington Post,* and is a guest lecturer in globalization and finance at The Johns Hopkins University, Georgetown University, Trinity College Dublin, the U.S. Army War College, the National Defense University, and the School of Advanced International Studies.